\documentclass[runningheads]{llncs}
\usepackage{graphicx}
\usepackage{subfig}
\usepackage{amsmath,amssymb} 
\usepackage{gensymb}
\usepackage{color}
\usepackage{multirow}
\usepackage[width=122mm,left=12mm,paperwidth=146mm,height=193mm,top=12mm,paperheight=217mm]{geometry}
\usepackage[pagebackref=true,breaklinks=true,letterpaper=true,colorlinks,bookmarks=false]{hyperref}
\usepackage{booktabs}
\usepackage[font=small,labelfont=bf,tableposition=top]{caption}
\DeclareCaptionLabelFormat{andtable}{#1~#2  \&  \tablename~\thetable}
\def\mbf#1{\mathbf{#1}}
\def\mit#1{\mathit{#1}}

\newcommand{\minisection}[1]{\vspace{0.04in} \noindent {\bf #1:} }

\begin{document}
\pagestyle{headings}
\mainmatter

\title{Do We Really Need to Collect Millions of Faces for Effective Face Recognition?} 

\author{Iacopo Masi$^1$, Anh Tu\~{a}n Tr\~{a}n$^1$, Jatuporn Toy Leksut$^1$,\\Tal Hassner$^{2,3}$ and G\'{e}rard Medioni$^1$}
\institute{$^1$Institute for Robotics and Intelligent Systems, USC, CA, USA\\$^2$Information Sciences Institute, USC, CA, USA\\$^3$The Open University of Israel, Israel}

\titlerunning{Do We Really Need to Collect Millions of Faces}
\authorrunning{I. Masi, A.T. Tran, J. T. Leksut, T. Hassner and G. Medioni}

\maketitle

\begin{abstract}
Face recognition capabilities have recently made extraordinary leaps. Though this progress is at least partially due to ballooning training set sizes -- huge numbers of face images downloaded and labeled for identity -- it is not clear if the formidable task of collecting so many images is truly necessary. We propose a far more accessible means of increasing training data sizes for face recognition systems. Rather than manually harvesting and labeling more faces, we simply synthesize them. We describe novel methods of enriching an existing dataset with important facial appearance variations by manipulating the faces it contains. We further apply this synthesis approach when matching query images represented using a standard convolutional neural network. The effect of training and testing with synthesized images is extensively tested on the LFW and IJB-A (verification and identification) benchmarks and Janus CS2. The performances obtained by our approach match state of the art results reported by systems trained on millions of downloaded images. 
\end{abstract}

\section{Introduction}\label{sec:intro}
The recent impact of deep Convolutional Neural Network (CNN) based methods on machine face recognition capabilities has been nothing short of revolutionary. The conditions under which faces can now be recognized and the numbers of faces which systems can now learn to identify improved to the point where some consider machines to be better than humans at this task. This remarkable advancement is partially due to the gradual improvement of new network designs which offer better performance. However, alongside developments in network architectures, it is also the underlying ability of CNNs to learn from massive training sets that allows these techniques to be so effective. 

Realizing that effective CNNs can be made even more effective by increasing their training data, many begun focusing efforts on harvesting and labeling large image collections to better train their networks. In~\cite{taigman2014deepface}, a standard CNN was trained by Facebook using 4.4 million labeled faces and shown to achieve what was, at the time, state of the art performance on the Labeled Faces in the Wild (LFW) benchmark~\cite{LFWTech}. Some time later,~\cite{parkhi2015deep} proposed the VGG-Face representation, trained on 2.6 million faces, and Face++ proposed its Megvii System~\cite{megaVii}, trained on 5 million faces. All, however, pale in comparison to the Google FaceNet~\cite{schroff2015facenet} which used 200 million labeled faces for its training. 

Making networks better by collecting and labeling huge training sets is, unfortunately, not an easy game to play. The effort required to download, process and label millions of images from the Internet with reliable subject names is daunting. To emphasize this, the bigger sets,~\cite{taigman2014deepface} and~\cite{schroff2015facenet}, required the efforts of large scale commercial organizations to assemble (Facebook and Google, resp.) and none of these sets was publicly released by its owners. By comparison, the largest face recognition training set which is publicly available is the CASIA WebFace collection~\cite{yi2014learning} weighing in at a mere 495K images, several orders of magnitudes smaller than the two bigger commercial sets\footnote{MegaFace~\cite{kemelmacher2016megaface} is larger than CASIA, but was designed as a testing set and so provides few images per subject. It was consequently never used for training CNN systems.}.

\begin{figure}[t]
\centering
\subfloat[Face set statistics]{
  \resizebox{0.53\textwidth}{!}{
  
    \begin{tabular}[b]{lccc}
    \toprule
    \textbf{Dataset} & \textbf{\#ID} & \textbf{\#Img} & \textbf{\#Img$\mathbin{/}$\#ID} \\ \hline
    Google~\cite{schroff2015facenet} & 8M & 200M  & 25  \\ 
    Facebook~\cite{taigman2014deepface}  & 4,030  &  4.4M  & 1K  \\  
    VGG Face~\cite{parkhi2015deep} & 2,622 &  2.6M & 1K  \\ 
	MegaFace~\cite{kemelmacher2016megaface} & 690,572 & 1.02M& 1.5\\
    CASIA~\cite{yi2014learning} & 10,575  & 494,414 & 46  \\ \hline
    Aug. pose+shape & 10,575 &  1,977,656 & 187 \\ 
    Aug. pose+shape+expr & 10,575 & 2,472,070 & 234  \\ 
    \bottomrule
    \end{tabular}
    }
  
  \label{tab:data}
}
\subfloat[Images for subjects]{
\raisebox{-.06\height}{
  \includegraphics[width=.44\textwidth]{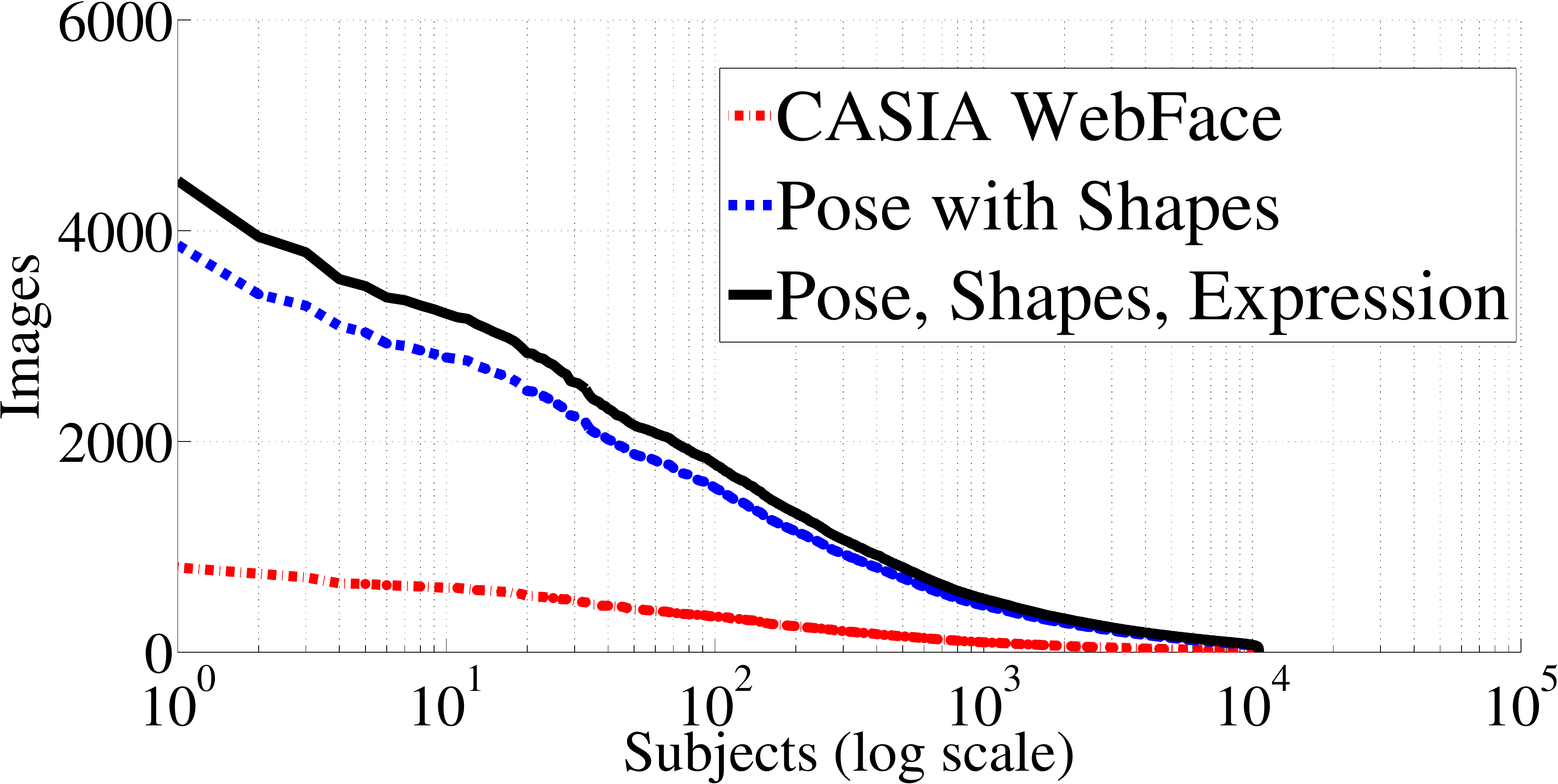}}
  \label{fig:distimg}
}
\caption{
(a) Comparison of our augmented dataset with other face datasets along with the average number of images per subject. (b) Our improvement by augmentation (Aug.) in the distribution of per-subject image numbers in order to avoid the long-tail effect of the CASIA set~\cite{yi2014learning} (also shown in the last two rows of (a)). 
}
\label{fig:intro}

\end{figure}

But downloading and labeling so many faces is more than just financially challenging. Fig.~\ref{tab:data} provides some statistical information on the larger face sets. Evidently, set sizes increase far faster than the numbers of images per-subject. This may imply that finding many images verified as belonging to the same subjects is difficult even when resources are abundant. Regardless of the reason, this is a serious problem: face recognition systems should learn to model not just inter-class appearance variations (differences between different people) but also intra-class variations (differences of appearance that do not change subject label) and so far this has been a challenge for data collection efforts.

In light of these challenges, it is natural to ask: is there no alternative to this labor intensive, data download and labeling approach to pushing recognition performance? Beyond potentially mitigating the challenges of data collection, this question touches a more fundamental issue. Namely, should images be processed with domain specific tools before being analyzed by CNNs, and if so, how?

In answer to these questions, we make the following contributions. {\bf (1)} We propose {\em synthesizing data} in addition to collecting it. We inflate the size of an existing training set, the CASIA WebFace collection~\cite{yi2014learning}, to several times its size using domain specific methods designed for face image synthesis (Fig.~\ref{fig:distimg}). We generate images which introduce new intra-class facial appearance variations, including pose (Sec.~\ref{sec:pose}), shape (Sec.~\ref{sec:3d}) and expression (Sec.~\ref{sec:expressions}). {\bf (2)} We describe a novel matching pipeline which uses similar synthesis methods at test time when processing query images. Finally, {\bf (3)}, we rigorously test our approach on the LFW~\cite{LFWTech}, IJB-A (verification and identification) and CS2 benchmarks~\cite{klare2015pushing}. Our results show that a CNN trained using these generated faces matches state of the art performance reported by systems trained on millions of faces downloaded from the web and manually processed\footnote{Code and data will be publicly available. Please see~\url{www.openu.ac.il/home/hassner/projects/augmented_faces} for updates.}.

We note that our approach can be considered a novel face data augmentation method (Sec.~\ref{sec:related}): A {\em domain specific} data augmentation approach. Curiously, despite the success of existing generic augmentation methods, we are unaware of previous reports of applying the easily accessible approach described here to generate face image training data, or indeed training for any other class.

\section{Related work}\label{sec:related}

\minisection{Face recognition} 
Face recognition is one of the central problems in computer vision and, as such, work on this problem is extensive. As with many other computer vision problems, face recognition performances sky rocketed with the introduction of deep learning techniques and in particular CNNs. Though CNNs have been used for face recognition as far back as~\cite{lawrence1997face}, only when massive amounts of data became available did their performance soar. This was originally demonstrated by the Facebook DeepFace system~\cite{taigman2014deepface}, which used an architecture not unlike the one used by~\cite{lawrence1997face}, but with over 4 million images used for training they obtained far more impressive results.

Since then, CNN based recognition systems continuously cross performance barriers with some notable examples including the Deep-ID 1-3 systems~\cite{sun2014deep,sun2014deep2,sun2015deepid3}. They and many others since, developed and trained their systems using far fewer training images, at the cost of somewhat more elaborate network architectures.

Though novel network architecture designs can lead to better performance, further improvement can be achieved by collecting more training data. This has been demonstrated by the Google FaceNet team~\cite{schroff2015facenet}, who developed and trained their system on 200 million images. Besides improving results, they also offered a fascinating analysis of the consequences of adding more data: apparently, there is a significant diminishing returns effect when training with increasing image numbers. Thus, the leap in performance obtained by going from thousands of images to millions is substantial but increasing the numbers further provides smaller and smaller benefits. One way to explain this is that the data they and others used suffers from a {\em long tail} phenomenon~\cite{yi2014learning}, where most subjects in these huge datasets have very few images available for the network to learn intra-subject appearance variations from.  

These methods were all evaluated on the LFW dataset, which has been for some time a {\em standard de facto} for measuring face recognition performances. Many of these LFW results, however, are already reaching near-perfect performances, suggesting that LFW is no longer a challenging benchmark for today's systems. Another relevant benchmark, also frequently used to report performances, is the YouTube Faces (YTF) set~\cite{wolf2011effective}. It contains unconstrained face videos rather than images, but it too is quickly being saturated. 

Recently, a new benchmark was released in order to again push machine face recognition capabilities, the Janus set~\cite{klare2015pushing}. It offers several novelties compared to existing sets, including template based, rather than image based, recognition and a mix of both images and videos. It is also tougher than previous collections. Not surprisingly, dominating performance on Janus are CNN methods such as~\cite{chen2015unconstrained}.

\minisection{Data augmentation} 
Data augmentation techniques are transformations applied to the images used for training or testing, but without altering their labels. Such methods are well known to improve the performance of CNN based methods and prevent overfitting~\cite{Chatfield14}. These methods, however, typically involved generic image processing operations which do not exploit knowledge of the underlying problem domain to synthesize new appearance variations. 

Popular augmentation methods include simple, geometric transformations such as {\em oversampling} (multiple, translated versions of the input image obtained by cropping at different offsets)~\cite{krizhevsky2012imagenet,LH:CVPRw15:age}, {\em mirroring} (horizontal flipping)~\cite{Chatfield14,yang2015mirror}, rotating~\cite{Xie2015Holistically} the images as well as various photometric transformations~\cite{krizhevsky2012imagenet,Simonyan2015very,eigen2015predicting}.

Surprisingly, despite being widely recognized as highly beneficial to the training of CNN systems, we are unaware of previous attempts to go beyond these simple transformations as we proposed doing. One notable exception is the recent work of~\cite{mclaughlin2015data} which proposes to augment training data for a person re-identification network by replacing image backgrounds. We propose a far more elaborate, yet easily accessible means of data augmentation.

Finally, we note that the recent work of~\cite{xie2015hyper} describes a so-called task-specific data augmentation method. They, as well as~\cite{xu2015augmenting}, do not synthesize new data as we propose to do here, but rather offer additional means of collecting images from the Internet to improve learning in fine grained recognition tasks. This is, of course, very different from our own approach.

\minisection{Face synthesis for face recognition} 
The idea that face images can be synthetically generated in order to aid face recognition systems is not new. To our knowledge, it was originally proposed in~\cite{hassner2013viewing} and then effectively used by~\cite{taigman2014deepface} and~\cite{hassner2015effective}. Contrary to us, they all produced frontal faces which are presumably better aligned and easier to compare. They did not use other transformations to generate new images (e.g., other poses, facial expressions). More importantly, their images were used to {\em reduce} appearance variability, whereas we propose the opposite: to dramatically {\em increase} it to improve both training and testing.

\section{Synthesizing faces}\label{sec:system}
In this section we detail our approach to augmenting a generic face dataset. We use the CASIA WebFace collection~\cite{yi2014learning}, enriching it with substantially more per-subject appearance variations, yet without changing subject labels or losing meaningful information. Specifically, we propose to generate (synthesize) new face images, by introducing the following face specific appearance variations:

\begin{enumerate}
\item {\bf Pose:} Simulating face image appearances across unseen 3D viewpoints.
\item {\bf Shape:} Producing facial appearances using different 3D generic face shapes.
\item {\bf Expression:} Specifically, simulating closed mouth expressions.
\end{enumerate}
As previously mentioned, (1) can be considered an extension of {\em frontalization} techniques~\cite{hassner2015effective} to multiple views. Conceptually, they rendered new views to reduce variability for better alignment whereas we do this to increase variability to better capture intra-subject appearance variations. Also noteworthy is that (2) explicitly contradicts previous assumptions on the importance of 3D facial shape in recognizing faces (e.g.,~\cite{taigman2014deepface}): Contrary to these claims -- that shape carries important subject related information -- we ignore these cues by rendering the same face using different underlying shapes. As we later show, this introduces subtle appearance variations which provide meaningful information at training, but by no means change perceived subject identities.

\subsection{Pose variations}\label{sec:pose}
In order to generate unseen viewpoints given a face image $\mbf{I}$, we use a technique similar to the frontalization proposed by~\cite{hassner2015effective}. We begin by applying the facial landmark detector from~\cite{lp:landmark}. Given these detected landmarks we estimate the six degrees of freedom pose for the face in $\mbf{I}$ using correspondences between the detected landmarks $\mbf{p}_i \in \mathbb{R}^{2}$ and points $\mbf{P}_i \doteq \mbf{S(i)} \in \mathbb{R}^{3}$, labeled on a 3D generic face model $\mbf{S}$. Here, $i$ indexes specific facial landmarks in $\mbf{I}$ and the 3D shape $\mbf{S}$. 

As mentioned earlier, we use CASIA faces for augmentation. These faces are roughly centered in their images, and so detecting face bounding boxes was unnecessary. Instead, we used a fixed bounding box determined once beforehand.

\begin{figure}[t]
\centering
\includegraphics[width=.9\textwidth]{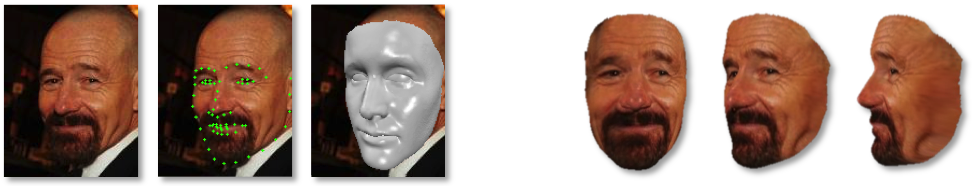}
\caption{Adding pose variations by synthesizing novel viewpoints. Left: Original image, detected landmarks, and 3D pose estimation. Right: rendered novel views.}
\label{fig:poses}
\vspace{-3mm}
\end{figure}

Given the corresponding landmarks $\mbf{p}_i \leftrightarrow \mbf{P}_i$ we use PnP~\cite{hartley2003multiple} to estimate extrinsic camera parameters, assuming the principal point is in the image center and then refining the focal length by minimizing landmark re-projection errors. This process gives us a perspective camera model mapping the generic 3D
shape $\mbf{S}$ on the image such as $\mbf{p}_i \sim \mbf{M}~\mbf{P}_i$ where $\mbf{M} = \mbf{K}~\left[ \mbf{R}~\mbf{t} \right]$ is the camera matrix. 

Given the estimated pose $\mbf{M}$, we decompose it to obtain a rotation matrix $\mbf{R} \in \mathbb{R}^{3 \times 3}$ containing rotation angles for the 3D head shape with respect to the image. We then create new rotation matrices $\mbf{R}_{\theta}' \in \mathbb{R}^{3 \times 3}$ for unseen (novel) viewpoints by sampling different yaw angles $\theta$. In particular, since CASIA images are biased towards frontal faces, given an image $\mbf{I}$ we render it at the fixed yaw values $\theta = \{ 0\degree, \pm40\degree, \pm75\degree \}$.
Rendering itself is derived from~\cite{hassner2015effective}, including soft-symmetry. Fig.~\ref{fig:poses} shows viewpoint (pose) synthesis results for a training subject in CASIA, illustrating the 3D pose estimation process.

Note that in practice, faces are rendered with a uniform black background not shown here (original background from the image was not preserved in rendering).

\begin{figure}[t]
\centering

\includegraphics[width=.98\textwidth]{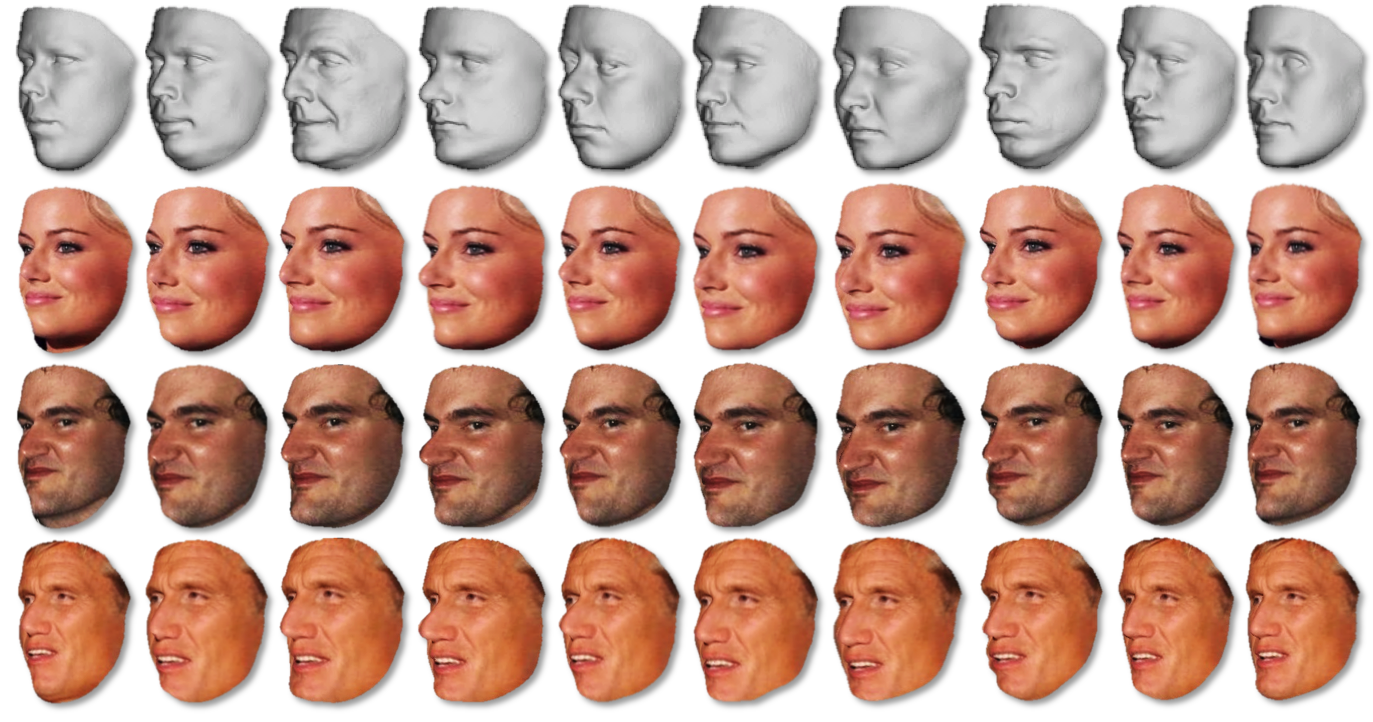} 
\caption{Top: The ten generic 3D face shapes used for rendering. Bottom: Faces rendered with the generic appearing right above them. Different shapes induce subtle appearance variations yet do not change the perceived identity of the face in the image.}
\label{fig:shapes}
\end{figure}

\subsection{3D shape variations}\label{sec:3d}
In the past, some argued that to truthfully capture the appearance of a subject's face under different viewpoints, its true 3D form must be used for the rendering. They therefore attempted to estimate this 3D shape from the image directly prior to frontalization~\cite{taigman2014deepface}. Because this reconstruction process is unstable, particularly for challenging, unconstrained images, Hassner \emph{et al.}~\cite{hassner2015effective} instead used a single generic 3D face to frontalize all face images. We propose the following simple compromise between these two approaches.

Rather than using a single generic 3D shape or estimating it from the image directly, we extend the procedure described in Sec.~\ref{sec:pose} to multiple generic 3D faces. In particular we add the set of generic 3D shapes $\mathcal{S} = \{ \mbf{S}_j \}_{j=1}^{10}$. We then simply repeat the pose synthesis procedure with these ten shapes rather than using only a single 3D shape. 

We used 3D generic shapes from the publicly available Basel 3D face set~\cite{bfm2009}. It includes 10 high quality 3D face scans captured from different people with different face shapes. The subjects vary in gender, age, and weight. The models are further well aligned to each other, hence requiring 3D landmarks to be selected only once, on one of these 3D faces, and then directly transferring them to the other nine models. Fig.~\ref{fig:shapes} shows the ten generic models used here, along with images rendered to near profile view using each of these shapes. Clearly, subjects in these images remain identifiable, despite the different underlying 3D shape, meeting the augmentation requirement of not changing subject labels. Yet each image is slightly but noticeably different from the rest, introducing  appearance variations to this subject's image set.

\subsection{Expression variations}\label{sec:expressions}
\begin{figure}[t]
\centering
\includegraphics[width=.95\textwidth]{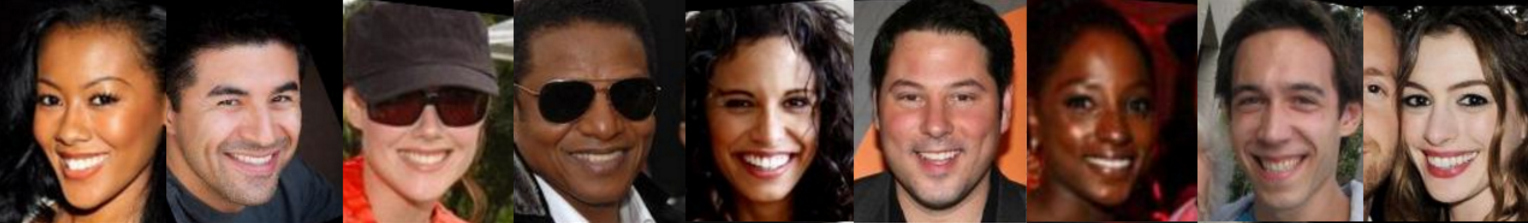}
\includegraphics[width=.95\textwidth]{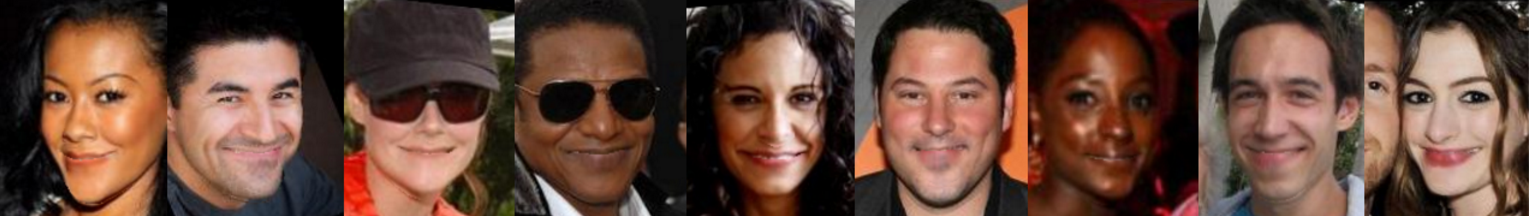}
\caption{Expression synthesis examples. Top: Example face images from the CASIA WebFace dataset. Bottom: Synthesized images with closed mouths.}
\label{fig:expr}
\end{figure}

In addition to pose and shape, we also synthesize expression variations, specifically reducing deformations around the mouth. Given a face image $\mbf{I}$ and its 2D detected landmarks $\mbf{p}_i$, and following pose estimation (Sec.~\ref{sec:pose}) we estimate facial expression by fitting a 3D expression {\em Blendshape}, similarly to~\cite{blendshapes}. 
This  is a linear combination of 3D generic face models with various basis expressions, including {\em mouth-closed}, {\em mouth-opened} and {\em smile}. 
Following alignment of the 3D face model and the 2D face image in both pose and expression, we perform image-based texture mapping to register the face texture onto the model. This is useful to quickly assign texture to our face model given that only one image is available. To synthesize expression, we manipulate the 3D textured face model to exhibit new expressions and render it back to the original image. This technique allows us to render a normalized expression where other image details, including hair and background, remain unchanged. In our experiments we do this to produce images with closed mouths. Some example synthesis results are provided in Fig.~\ref{fig:expr}. Though slight artifacts are sometimes introduced by this process (some can be seen in Fig.~\ref{fig:expr}) these typically do not alter the general facial appearances and are less pronounced than the noise often present in unconstrained images.

\section{Face recognition pipeline}\label{sec:cnn}
Data augmentation techniques are not restricted to training and are often also applied at test time. Our augmentations provide opportunities to modify the matching process by using different augmented versions of the input image. We next describe our recognition pipeline including these and other novel aspects. 

\subsection{CNN training with our augmented data}\label{sec:training}

\minisection{Augmented training data}
Our pipeline employs a single CNN trained on both real and augmented data generated as described in Sec.~\ref{sec:system}. Specifically, training data is produced from the original CASIA WebFace images. It consists of the following types of images: (i) original CASIA images following alignment by a simple, in-plane similarity transform to two coordinate systems: roughly frontal facing faces (face yaw estimates in $[ -30\degree...~30\degree ]$) are aligned using nine landmarks on an ideal frontal template, while profile images (all other yaw angles) are aligned using the visible eye and the tip of the nose. (ii) Each image in CASIA is rendered from three novel views in yaw angles $\{0\degree,\pm40\degree,\pm75\degree\}$, as described in Sec.~\ref{sec:pose}. (iii) Synthesized views are produced by randomly selecting a 3D generic face model from $\mathcal{S}$ as the underlying face shape (see Sec.~\ref{sec:3d}), thereby adding shape variations. (iv) Finally, a mouth neutralized version of each image is also added to the training (Sec.~\ref{sec:expressions}). This process raises the total number of images available for training from 494,414 in the original CASIA WebFace set to a total of 2,472,070 images in our complete (pose+shape+expression) augmented dataset. Note that this process leaves the number of CASIA WebFace subjects unchanged, inflating only the number of images per subject (Fig.~\ref{fig:distimg}).

\minisection{CNN fine-tuning}\label{sec:finetune}
We use the very deep VGGNet~\cite{Simonyan2015very} CNN with 19 layers, trained on the large scale image recognition benchmark (ILSVRC)~\cite{russakovsky2014imagenet}. We fine tune this network using our augmented data. To this end, we keep all layers $\{\mbf{W}_k,\mbf{b}_k\}_{k=1}^{19}$ of VGGNet except for the last linear layer (FC8) which we train from scratch. This layer produces a mapping from the embedded feature $\mbf{x} \in \mathbb{R}^{D}$ (FC7) to the subject labels $N=10,575$ of the augmented dataset. It  computes $\mbf{y}=\mbf{W}_{19}\mbf{x}+\mbf{b}_{19}$, where $\mbf{y} \in \mathbb{R}^{N}$ is the linear response of FC8. Fine-tuning is performed by minimizing the soft-max loss:
\begin{equation}
\mathcal{L}(\{\mbf{W}_k,\mbf{b}_k\}) = - \sum_{t}\ln \bigg( \frac{ \mit{e}^{\mbf{y}_{l}} }{\sum_{g=1}^{N}\mit{e}^{\mbf{y}_g} } \bigg) 
\label{eq:loss}
\end{equation}
where $l$ is the ground-truth index over $N$ subjects and $t$ indexes all training images. Eq.~(\ref{eq:loss}) is optimized using Stochastic Gradient Descent (SGC) with standard L2 norm over the learned weights. When performing back-propagation, we learn FC8 faster since it is trained from scratch while other network weights are updated with a learning rate an order of magnitude lower than FC8. 

Specifically, we initialize FC8 with parameters drawn from a Gaussian distribution with zero mean and standard deviation 0.01. Bias is initialized with zero. The overall learning rate $\mu$ for the entire CNN is set to $0.001$, except FC8 which uses learning rate of $\mu\times10$. We decrease learning rate by an order of magnitude once validation accuracy for the fine tuned network saturates.  Meanwhile, biases are learned twice as fast as the other weights. For all the other parameter settings we use the same values as originally described in~\cite{krizhevsky2012imagenet}.

\subsection{Face recognition with synthesized faces}\label{sec:matching}

\minisection{General matching process}\label{sec:generalMatch}
After training the CNN, we use the embedded feature vector $\mbf{x} = f(\mbf{I};\{\mbf{W}_k,\mbf{b}_k\} )$ 
from each image $\mbf{I}$ as a face representation. Given two input images $\mbf{I}_p$ and $\mbf{I}_q$, their similarity, $s(\mbf{x}_p, \mbf{x}_q)$ is simply the normalized cross correlation (NCC) of their feature vectors.

The value $s(\mbf{x}_p, \mbf{x}_q)$ is the recognition score at the image level. In some cases a subject is represented by multiple images (e.g., a {\em template}, as in the Janus benchmark~\cite{klare2015pushing}). This plurality of images can be exploited to improve recognition at test time. In such cases, image sets are defined by $\mathcal{P}=\{\mbf{x}_1,...,\mbf{x}_P \}$ and $\mathcal{Q}=\{\mbf{x}_1,...,\mbf{x}_Q \}$ and a similarity score is defined between them: $s(\mathcal{P},\mathcal{Q})$.

Specifically, we compute the pair-wise image level similarity scores, $s(\mbf{x}_p, \mbf{x}_q)$, for all $\mbf{x}_p\in \mathcal{P}$ and $\mbf{x}_q\in\mathcal{Q}$, and pool these scores using a SoftMax operator, $s_{\beta}(\mathcal{P},\mathcal{Q})$ (Eq.(\ref{eq:softmax}), below). Though the use of SoftMax here is inspired by the SoftMax loss often used by CNNs, its aim is to get a robust score regression instead of a distribution over the subjects. SoftMax for set fusion can be seen as a weighted average in which the weight depends on the score when performing recognition. It is interesting to note here that the SoftMax hyper-parameter $\beta$ controls the trade-off between averaging the scores or taking the maximum (or minimum). That is:
\begin{equation}
\mbox{s}_{\beta}(\cdot,\cdot)   = \begin{cases} \max(\cdot) &  \mbox{if } \beta \to  \infty \\ 
\mbox{avg}(\cdot) & \mbox{if } \beta=0\\
\min(\cdot) & \mbox{if } \beta \to  -\infty \end{cases}
\text{and}~ 
s_{\beta}(\mathcal{P}, \mathcal{Q}) = \frac{\sum_{p \in \mathcal{P}, q \in \mathcal{Q}} s(\mbf{x}_p, \mbf{x}_q)\mit{e}^{\beta ~ s(\mbf{x}_p, \mbf{x}_q)} }{\sum_{p \in \mathcal{P}, q \in \mathcal{Q}} \mit{e}^{\beta ~ s(\mbf{x}_p, \mbf{x}_q)}}.
\label{eq:softmax}
\end{equation}
Pair-wise scores are pooled using Eq.(\ref{eq:softmax}) and we finally average the SoftMax responses over multiple values of $\beta= \left [ 0 ... 20 \right ]$ to get the final similarity score:
\begin{equation}
s(\mathcal{P}, \mathcal{Q}) = \frac{1}{21}\sum_{\beta=0}^{20} s_{\beta}(\mathcal{P}, \mathcal{Q}).
\end{equation}
The use of positive values for $\beta$ is motivated by the fact what we are using a score for recognition, so the higher the value, the better. In our experiments we found that the SoftMax operator reaches a remarkable trade-off between averaging the scores and taking the maximum. The improvement given by the proposed SoftMax fusion is shown in Tab.~\ref{tab:softmax}: we can see how the proposed method largely outperforms standard fusion techniques on IJB-A, in which subjects are described by templates.

\begin{table}

\centering

\begin{tabular}{l||c|c||c|c|c}

\toprule

\textbf{Fusion$\downarrow$} &  \multicolumn{2}{c||}{\textbf{IJB-A Ver. (TAR)}} & \multicolumn{3}{c}{\textbf{IJB-A Id. (Rec. Rate)}}  \\ \hline 
 
 Metrics $\rightarrow$ & FAR0.01& FAR0.001& Rank-1  & Rank-5  & Rank-10 \\ \hline 

Min   & 26.3 & 11.2 & 33.1 & 56.1 & 66.8  \\ 

Max & 77.6  & 46.4 & 84.8 & 93.3 & 95.6  \\  

Mean & 79.9 & 53.0 & 84.6 &94.7 & 96.6  \\ \hline 
SoftMax & \textbf{86.6} & \textbf{63.6} & \textbf{87.2} & \textbf{94.9} & \textbf{96.9}  \\ 

\bottomrule

\end{tabular}

\vspace{-1mm}
\caption{SoftMax template fusion for score pooling vs. other standard fusion techniques on the IJB-A benchmark for verification (ROC) and identification (CMC) resp.}
\label{tab:softmax}
\vspace{-4mm}
\end{table}

\minisection{Exploiting pose augmentation at test time}\label{sec:viewbased}
The Achilles heel of many face recognition systems is cross pose face matching; particularly when one of the two images is viewed at an extreme, near profile angle~\cite{yi2013towards,li2013probabilistic,ding2015multi}. Directly matching two images viewed from extremely different viewpoints often leads to poor accuracy as the difference in viewpoints affects the similarity more than subject identities. To mitigate this problem, we suggest rendering both images from the same view: one that is close enough to the viewpoint of both images. To this end, we leverage our pose synthesis method of Sec.~\ref{sec:pose} to produce images in poses better suited for recognition and matching.

Cross pose rendering can, however, come at a price: Synthesizing novel views for faces runs the risk of producing meaningless images whenever facial landmarks are not accurately localized and the pose estimate is wrong. Even if pose was correctly estimated, warping images across poses involves interpolating intensities, which leads to smoothing artifacts and information loss. Though this may affect training, it is far more serious at test time where we have few images to compare and ruining one or both can directly affect recognition accuracy.

Rather than commit to pose synthesis or its standard alternative, simple yet robust in-plane alignment, we propose to use both: We found that pose synthesis and in-plane alignment are complimentary and by combining the two alignment techniques recognition performance improves. For an image pair $(\mbf{I}_p, \mbf{I}_q)$ we compute two similarity scores. One score is produced using in-plane aligned images and the other using images rendered to a mutually convenient view. This view is determined as follows: If the two images are near frontal then we render them to frontal view (essentially frontalizing them~\cite{hassner2015effective}), if they are both near profile we render to $75\degree$, otherwise we render both to $40\degree$.

When matching templates (image sets), $(\mathcal{P}, \mathcal{Q})$, scores computed for in-plane aligned image pairs and pose synthesized pairs are pooled separately using Eq.~(\ref{eq:softmax}). This is equivalent to comparing the two sets $\mathcal{P}$ and $\mathcal{Q}$ twice, once for each alignment method. These two similarities are then averaged for the final template level score.

\section{Experiments}\label{sec:experiments}
We tested our approach extensively on the recently released IARPA Janus benchmarks~\cite{klare2015pushing} and LFW~\cite{LFWTech}. We perform a minimum of database specific  training, using the training images prescribed by each benchmark protocol. Specifically, we perform Principal Component Analysis (PCA) on the training images of the target dataset with the features $\mbf{x}$ extracted from the CNN trained on augmented data. This did not include dimensionality reduction; we did not cut any component after PCA projection. Following this, we apply root normalization to the new projected feature, i.e., $\mbf{x} \rightarrow  \mbf{x}^c$, as previously proposed for the Fisher Vector encoding in~\cite{sanchez2013image}. We found that a value of $c=0.65$ provides a good baseline across all the experiments.
For each dataset we report the contribution of each augmentation technique compared with state-of-the-art methods which use millions of images to train their deep models.
\begin{figure}[t]
\centering
\subfloat[Contribution of augmentation]{
\includegraphics[width=.23\textwidth]{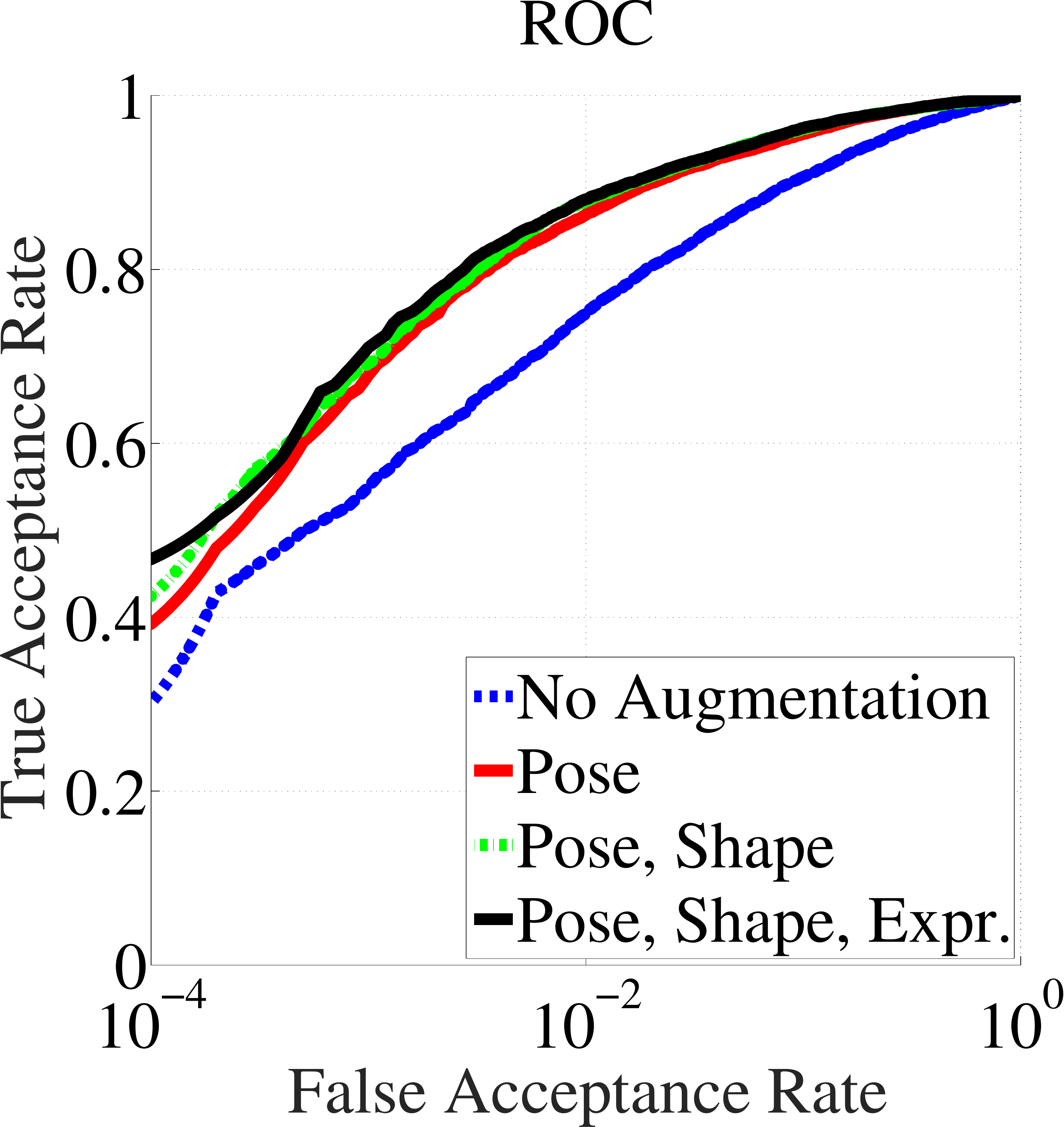}
\includegraphics[width=.24\textwidth]{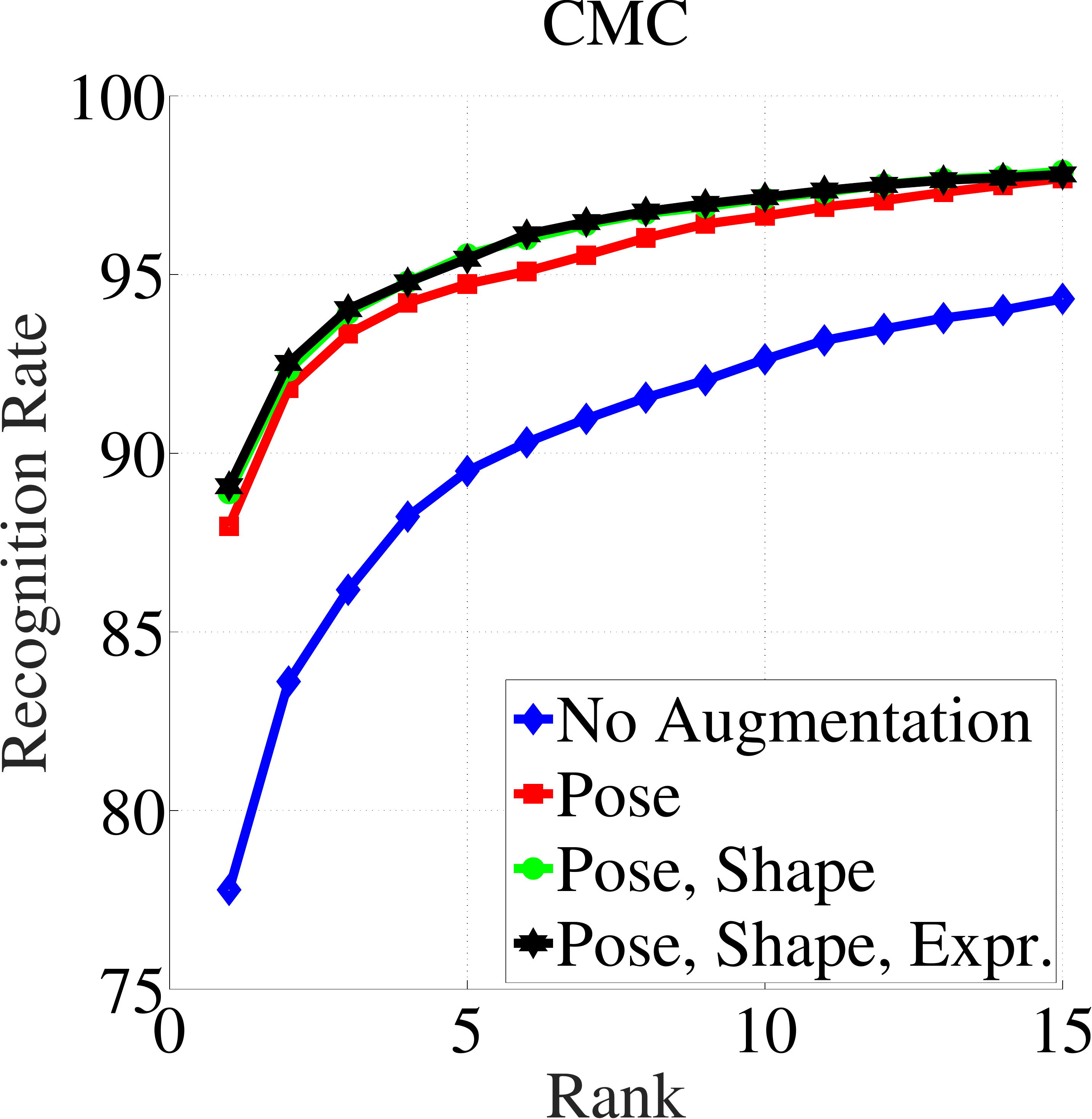}
\label{fig:ijba-ablation1}
}
\subfloat[Matching methods]{
\includegraphics[width=.23\textwidth]{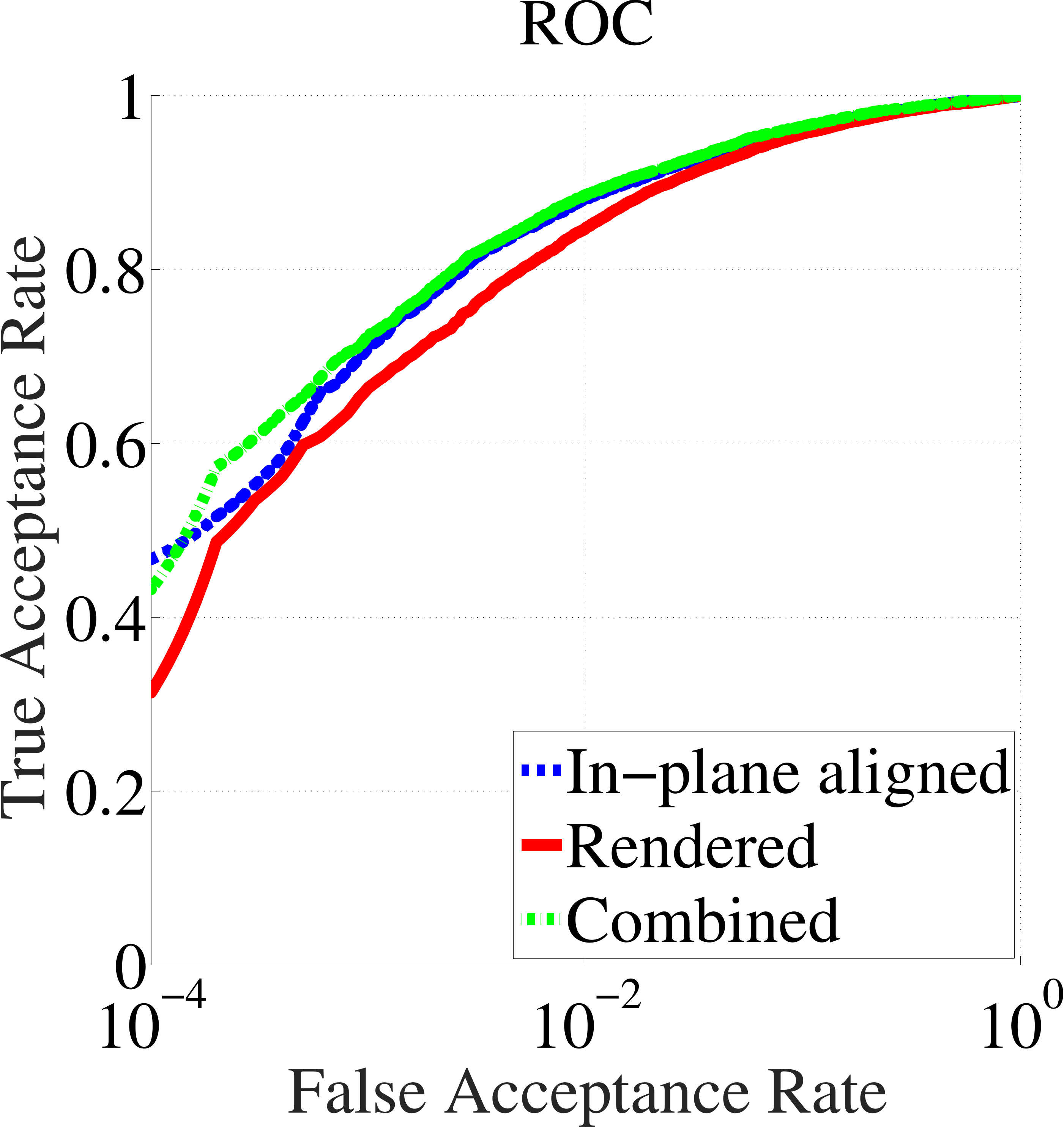}
\includegraphics[width=.24\textwidth]{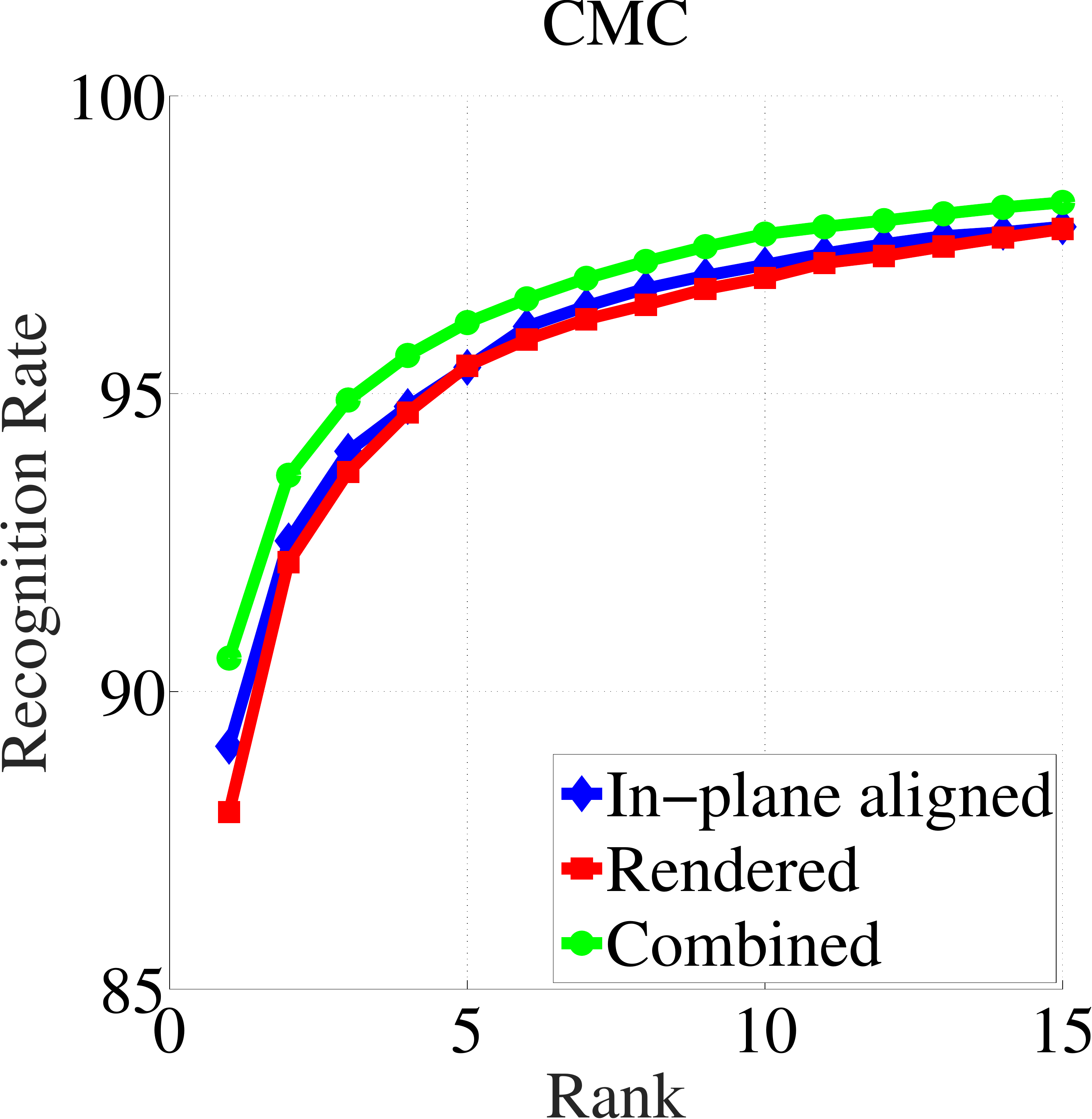}
\label{fig:ijba-ablation2}
}
\caption{Ablation study of our data synthesis and test time matching methods on IJB-A.}
\label{fig:ijba-ablation}
\vspace{-3mm}
\end{figure}
\subsection{Results on the IJB-A benchmarks}\label{sec:janus}
IJB-A is a new publicly available benchmark released by NIST\footnote{IJB-A data and splits are available under request at~\url{http://www.nist.gov/itl/iad/ig/facechallenges.cfm}} to raise the challenges of unconstrained face identification and verification methods. Both IJB-A and the Janus CS2 benchmark share the same subject identities, represented by images viewed in extreme conditions, including pose, expression and illumination variations, with IJB-A splits generally considered more difficult than those in CS2. The IJB-A benchmarks consist of face verification (1:1) and face identification (1:N) tests. Contrary to LFW, Janus subjects are described using templates containing mixtures of still-images and video frames. 

It is important to note that the Janus set has some overlap with the images in the CASIA WebFace collection. In order to provide fair comparisons, our CNNs were fine tuned on CASIA subjects that are {\em not} included in Janus (Sec.~\ref{sec:finetune}).

\minisection{Face detections} Our pipeline uses the facial landmark detector of~\cite{lp:landmark} for head pose estimation and alignment. Although we found this detector quite robust, it failed to detect landmarks on some of the more challenging Janus faces. Whenever the detector failed on all the images in the same template, we use the images cropped to their facial bounding boxes as provided in the Janus data. 

\minisection{Video pooling} We note that whenever face templates include multiple frames from a single video, we pool together CNN features extracted from the same video: this, by simple element wise average over all the features extracted from that video's frames. We emphasize that features are not pooled across videos but only within each video. Similar pooling techniques were very recently demonstrated to provide substantial performance enhancements (e.g.,~\cite{su2015multi}) but, to our knowledge, never for faces or in the manner suggested here. We refer to this technique as {\em video pooling} and report its influence on our system, and, whenever possible, for our baselines. 

In all our IJB-A and Janus CS2 results this method provided noticeable performance boosts: we compare video pooling to pair-wise single image comparisons (referred as {\em without video pooling} in our results).

\begin{table*}[t]

\centering

\resizebox{\textwidth}{!}{

\def\arraystretch{1.4}

\begin{tabular}{l||c|c||c|c|c||c|c||c|c|c}

\hline
 &  \multicolumn{5}{c||}{\textbf{Without Video pooling}} & \multicolumn{5}{c}{\textbf{With Video pooling}} \\ \cline{1-11}
\textbf{Augmentation $\downarrow$} &  \multicolumn{2}{c||}{\textbf{IJB-A Ver. (TAR)}} & \multicolumn{3}{c||}{\textbf{IJB-A Id. (Rec. Rate)}} & \multicolumn{2}{c||}{\textbf{IJB-A Ver. (TAR)}}  & \multicolumn{3}{c}{\textbf{IJB-A Id. (Rec. Rate)}} \\ \hline

  Metrics $\rightarrow$ & FAR0.01& FAR0.001& Rank-1  & Rank-5  & Rank-10  & FAR0.01 & FAR0.001  & Rank-1 &   Rank-5 &   Rank-10  \\ \cline{1-11}
\hline

No Augmentation   & 74.5 & 54.3 & 77.1  & 89.0 & 92.3 & 75.0 & 55.0  & 77.8 & 89.5 & 92.6  \\
Pose   & 84.9 & 62.3 & 86.3 & 94.5 & 96.5 &  86.3   &  67.9  & 88.0  & 94.7  &   96.6 \\ 

Pose, Shapes & 86.3 & 62.0 & 87.0  & 94.8  & \textbf{96.9} & 87.8 &  69.2 & 88.9  &  \textbf{95.6}  &   97.1    \\ \hline 

 Pose, Shapes, Expr. & \textbf{86.6}  & \textbf{63.6} & \textbf{87.2} & \textbf{94.9} & \textbf{96.9} & \textbf{88.1}  &  \textbf{71.0}  & \textbf{89.1} &  95.4   &   \textbf{97.2}  \\ 

\hline

\end{tabular}

}

\caption{Effect of each augmentation on IJB-A performance on verification (ROC) and identification (CMC), resp. Only in-plane aligned images used in these tests.}

\label{tab:1}

\end{table*}

\begin{table*}

\centering

\resizebox{\textwidth}{!}{

\def\arraystretch{1.5}

\begin{tabular}{l||c|c||c|c|c||c|c||c|c|c}

\hline

 &  \multicolumn{5}{c||}{\textbf{Without Video pooling}} & \multicolumn{5}{c}{\textbf{With Video pooling}} \\ \cline{1-11}
\textbf{Image type $\downarrow$} &  \multicolumn{2}{c||}{\textbf{IJB-A Ver. (TAR)}} & \multicolumn{3}{c||}{\textbf{IJB-A Id. (Rec. Rate)}} & \multicolumn{2}{c||}{\textbf{IJB-A Ver. (TAR)}}  & \multicolumn{3}{c}{\textbf{IJB-A Id. (Rec. Rate)}} \\ \cline{1-11}

  Metrics $\rightarrow$ & FAR0.01& FAR0.001& Rank-1  & Rank-5  & Rank-10  & FAR0.01 & FAR0.001  & Rank-1 &   Rank-5 &   Rank-10  \\ \cline{1-11}

In-plane aligned   & 86.6 & 63.6& 87.2 & 94.9 & 96.9  & 88.1  &  71.0  & 89.1  &  95.4   &   97.2\\ 

Rendered & 84.7  & 64.6 & 87.3 & 95.0 & 96.8 & 84.8   &  66.4 & 88.0  &  95.5   &   96.9    \\ \hline

Combined & \textbf{87.8} & \textbf{67.4} & \textbf{89.5} & \textbf{95.8} & \textbf{97.4}  & \textbf{88.6}   &  \textbf{72.5} & \textbf{90.6}  &  \textbf{96.2}    &   \textbf{97.7}  \\ 

\hline

\end{tabular}

}
\caption{Effect of in-plane alignment and pose synthesis at test-time (matching) on IJB-A dataset respectively for verification (ROC) and identification (CMC).}
\label{tab:2}
\vspace{-2mm}
\end{table*}

\minisection{Ablation Study}
We provide a detailed analysis of each augmentation technique on the challenging IJB-A dataset. Clearly, the biggest contribution is given by pose augmentation (red curve) over the baseline (blue curve) in Fig.~\ref{fig:ijba-ablation1}. The improvement is especially noticeable in the rank-1 recognition rate for the identification protocol. The effect of video pooling along with each data augmentation method is provided in Tab.~\ref{tab:1}. 

We next evaluate the effect of pose synthesis at test time combined with the standard in-plane alignment (Sec.~\ref{sec:viewbased}), in Tab~\ref{tab:2} and in Fig.~\ref{fig:ijba-ablation2}. Evidently, these methods combined contribute to achieving state-of-the-art performance on the IJB-A benchmark. We conjecture that this is mainly due to three contributions: domain-specific augmentation when training the CNN, combination of SoftMax operator, video pooling and finally pose synthesis at test time.

\minisection{Comparison with the state-of-the-art}
Our proposed method achieves state of the art results in the IJB-A benchmark and Janus CS2 dataset. In particular, it largely improves over the off the shelf commercial systems COTS and GOTS~\cite{klare2015pushing} and Fisher Vector encoding using frontalization~\cite{umd:FV}. This gap can be explained by the use of deep learning alone. Even compared with deep learning based methods, however, our approach achieves superior performance and with very wide margins. This is true even comparing our results to~\cite{80Msearch}, who use seven networks and fuse their output with the COTS system. Moreover, our method improves in IJB-A verification over~\cite{80Msearch}
in 15\% TAR at FAR=0.01 and $\sim$20\% TAR at FAR=0.001, also showing a better rank-1 recognition rate. 

It is interesting to compare our results to those reported by~\cite{chen2015unconstrained} and~\cite{Swami:UMD}. Both fine tuned their deep networks on the ten training splits of each benchmark, at substantial computational costs. Some idea of the impact this fine tuning can have on performance is available by considering the huge performance gap between results reported before and after fine tuning in~\cite{chen2015unconstrained}\footnote{The results reported in~\cite{chen2015unconstrained} with fine tuning on the training sets include system components not evaluated without fine tuning.}. Our own results, obtained by training our CNN once on augmented data, far outperform those of~\cite{Swami:UMD} also largely outperforming those reported by~\cite{chen2015unconstrained}. We conjecture that by training the CNN with augmented data we avoid further specializing all the parameters of the network on the target dataset. Tuning deep models on in-domain data is computationally expensive and thus, avoiding overfitting the network at training time is preferable.

\begin{table}[t]
\centering
\resizebox{\textwidth}{!}{
\def\arraystretch{1.2}
\begin{tabular}{l||c|c||c|c|c||c|c||c|c|c}
\hline
\textbf{Methods $\downarrow$} &  \multicolumn{2}{c||}{\textbf{JANUS CS2 Ver. (TAR)}} & \multicolumn{3}{c||}{\textbf{JANUS CS2 Id. (Rec. Rate)}} & \multicolumn{2}{c||}{\textbf{IJB-A Ver. (TAR)}}  & \multicolumn{3}{c}{\textbf{IJB-A Id. (Rec. Rate)}} \\ \cline{1-11}
  Metrics $\rightarrow$ & FAR0.01& FAR0.001& Rank-1  & Rank-5  & Rank-10  & FAR0.01 & FAR0.001  & Rank-1 &   Rank-5 &   Rank-10  \\ \cline{1-11}

COTS~\cite{klare2015pushing}   & 58.1 & 37 &55.1 & 69.4 & 74.1 & --   &   --  & --  & --   &   --    \\ 
GOTS~\cite{klare2015pushing}   & 46.7 & 25 &41.3 & 57.1 & 62.4 & 40.6 &  19.8 & 44.3  & 59.5 &   --    \\ 
OpenBR~\cite{openbr} & --  & --  & --  & --    & -- & 23.6  &  10.4 & 24.6 & 37.5 &   --    \\ 

Fisher Vector~\cite{umd:FV} & 41.1 & 25 & 38.1 & 55.9 & 63.7  & --   &   --  & --  & --   &   --  \\ 
Wang \emph{et al.}~\cite{80Msearch}  & --  & -- &  -- & --  & --  &  73.2  &  51.4 & 82.0 & 92.9 &  --  \\
Chen \emph{et al.}~\cite{chen2015unconstrained}  & 64.9 & 45 &  69.4 & 80.9 & 85.0 &  57.3 &   --  & 72.6  & 84.0 &  88.4  \\
Deep Multi-Pose~\cite{AbdAlmageed2016multipose} & 89.7 &  -- & 86.5 &   93.4 &  94.9 & 78.7    & --  & 84.6 & 92.7 &   94.7   \\  \hline
Chen \emph{et al.}~\cite{chen2015unconstrained} (f.t.) & 92.1 & 78 &  89.1 & \textbf{95.7} & \textbf{97.2} &  83.8 &   --  & 90.3  & \textbf{96.5}   &  \textbf{97.7}  \\
Swami S. \emph{et al.}~\cite{Swami:UMD} (f.t.) & -- &  -- & --  & --  & --  &  79   & 59  & 88 & 95 &   --   \\ \hline
Ours &  \textbf{92.6}  &  \textbf{82.4} &  \textbf{89.8} & {95.6}  & 96.9  & \textbf{88.6}   &  \textbf{72.5}  & \textbf{90.6}  & 96.2  &  \textbf{97.7} \\ 
\hline
\end{tabular}
}
\caption{Comparative performance analysis on JANUS CS2 and IJB-A respectively for verification (ROC) and identification (CMC). f.t. denotes fine tuning a deep network multiple times for each training split. A network trained once with our augmented data achieves mostly superior results, without this effort.}
\vspace{-3mm}
\label{tab:3}
\end{table}

\subsection{Results on Labeled Faces in the Wild}\label{sec:lfw}
For many years LFW~\cite{LFWTech} was the standard benchmark for unconstrained face verification. Recent methods dominating LFW scores use millions of images collected and labeled by hand in order to obtain their remarkable performances. To test our approach, we follow the standard protocol for unrestricted, labeled outside data and report the mean classification accuracy as well as the 100\% - EER (Equal Error Rate). We prefer to use 100\% - EER in general because it is not dependent on the selected classification threshold but we still report verification accuracy to be comparable with the other methods.

\begin{figure}[t]
\centering
\subfloat[Ablation Study]{
\includegraphics[width=.295\textwidth]{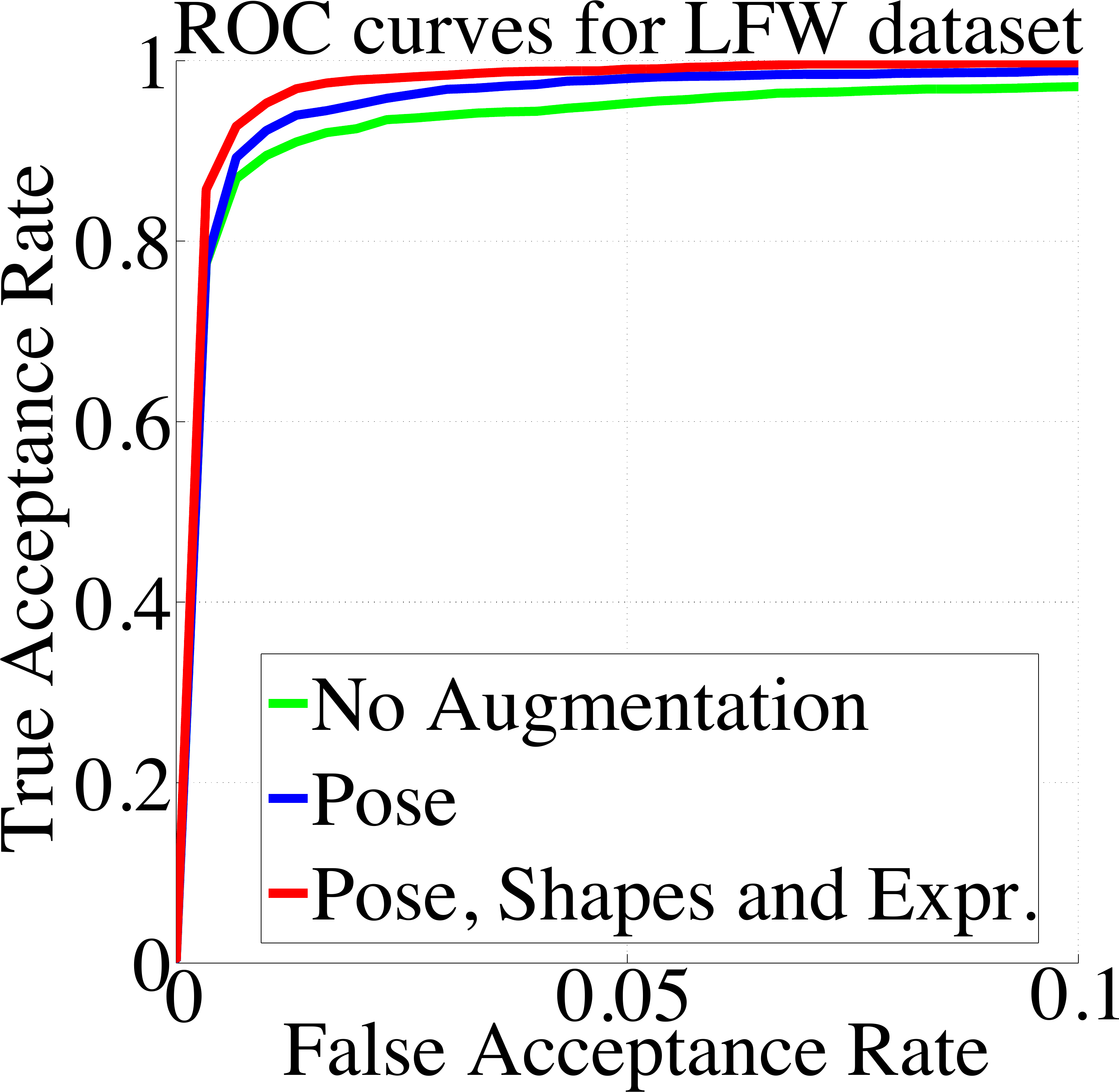}
\label{fig:lfw1}
}
\subfloat[Results for methods trained on millions of images]{
\resizebox{0.68\textwidth}{!}{
\begin{tabular}[b]{lccccc}
\toprule
\textbf{Method}& \textbf{Real} & \textbf{Synth} &\textbf{Net}& \textbf{Acc. (\%)} &\textbf{100\% - EER} \\ \hline
Fisher Vector Faces~\cite{Parkhi_2014_CVPR} & -- & --& -- & 93.0 & 93.1 \\ 
DeepFace~\cite{taigman2014deepface} & 4M & -- & 3 & 97.35 & -- \\ 
Fusion~\cite{Taigman_2015_CVPR} & 500M & -- & 5 & 98.37 & -- \\  
FaceNet~\cite{schroff2015facenet} & 200M &  -- &1 & 98.87 & -- \\ 
FaceNet + Alignment~\cite{schroff2015facenet} & 200M & -- & 1 & 99.63 & -- \\
VGG Face~\cite{parkhi2015deep} & 2.6M & -- & 1 & -- &  97.27 \\ 
VGG Face (triplet loss)~\cite{parkhi2015deep} & 2.6M & -- & 1 & 98.95 & 99.13 \\ \hline 
Us, no aug. &	495K & -- &1 &	95.31 &	95.26 \\
Us, aug. pose 	& 495K & 2M &	1	 & 97.01 &	96.93\\
Us, aug. pose, shape, expr. &	495K & 2.4M &	1	& 98.06	& 98.00\\
\bottomrule
\end{tabular}
}
\vspace{-3mm}
\label{fig:lfw2}
}
\caption{LFW verification results. (a) Break-down of the influence of different training data augmentation methods. (b)  Performance comparison with state of the art methods, showing the numbers of real (original) and synthesized training images, number of CNNs used by each system, accuracy and 100\%-EER.}
\label{fig:lfw}
\vspace{-3mm}
\end{figure}

\minisection{Improvement for each augmentation}
Fig.~\ref{fig:lfw1} provides ROC curves for each augmentation technique used in our approach. The green curve represents our baseline, that is the CNN trained on in-plane aligned images with respect to a frontal template. The ROC improves by a good margin when we inject unseen rendered images across poses into each subject. Indeed the 100\% - EER improves by +1.67\%. Moreover, by adding both shapes and expressions, performance improves even more, reaching 100\% - EER rate of 98.00\% (red curve).  See Tab.~\ref{fig:lfw2} for a comparison with methods trained on millions of downloaded images.

\subsection{Result summary}\label{sec:summary}
It is not easy to compare our results with those reported by others using millions of training images: Their system designs and implementation details are different from our own  and it is difficult to assess how different system components contribute to their overall performance. In particular, the reluctance of commercial groups to release their code and data makes it hard to say exactly how much performance our augmentation buys in comparison to their harvesting and curating millions of face images. 

Nevertheless, the results throughout this section clearly show that synthesizing training images using domain tools and knowledge leads to dramatic increase in recognition accuracy. This may be attributed to the potential of domain specific augmentation to infuse training data with important intra-subject appearance variations; the very variations that seem  hardest to obtain by simply downloading more images. As a bonus, it is a more accessible means of increasing training set sizes than downloading and labeling millions of additional faces. 

Finally, a comparison of our results on LFW to those reported by methods trained on millions of images (\cite{taigman2014deepface,Taigman_2015_CVPR,parkhi2015deep} and ~\cite{schroff2015facenet}), shows that with the initial set of less than 500K publicly available images from~\cite{yi2014learning}, our method surpasses those of~\cite{taigman2014deepface} and~\cite{parkhi2015deep} (without their metric learning, which was not applied here), falling only slightly behind the rest.

\section{Conclusions}\label{sec:conclusions}
This paper makes several important contributions. First, we show how domain specific data augmentation can be used to generate (synthesize) valuable additional data to train effective face recognition systems, as an alternative to expensive data collection and labeling. Second, we describe a face recognition pipeline with several novel details. In particular, its use of our data augmentation for matching across poses in a natural manner. Finally, {\em in answer to the question in the title}, our extensive analysis shows that though there is certainly a benefit to downloading increasingly larger training sets, much of this effort can be substituted by simply synthesizing more face images.

There are several compelling directions of extending this work. Primarily, the underlying idea of domain specific data augmentation can be extended in more ways (more facial transformations) to provide additional intra subject appearance variations. Appealing potential augmentation techniques, not used here, are facial age synthesis~\cite{kemelmacher2014illumination} or facial hair manipulations~\cite{nguyen2008image}. Finally, beyond faces there may be other domains where such approach is relevant and where the introduction of synthetically generated training data can help mitigate the many problems of data collection for CNN training.

\section*{Acknowledgments}
The authors wish to thank Jongmoo Choi for all his help in this project. This research is based upon work supported in part by the Office of the Director of National Intelligence (ODNI), Intelligence Advanced Research Projects Activity (IARPA), via IARPA 2014-14071600011. The views and conclusions contained herein are those of the authors and should not be interpreted as necessarily representing the official policies or endorsements, either expressed or implied, of ODNI, IARPA, or the U.S. Government.  The U.S. Government is authorized to reproduce and distribute reprints for Governmental purpose notwithstanding any copyright annotation thereon. Moreover, we gratefully acknowledge the support of NVIDIA Corporation with the donation of the NVIDIA Titan X GPU used for this research.

\bibliographystyle{ieee}

\end{document}